%% file: acl_latex.tex
\pdfoutput=1

\documentclass[11pt]{article}

\usepackage[final]{acl}

\usepackage{times}
\usepackage{latexsym}

\usepackage[T1]{fontenc}

\usepackage[utf8]{inputenc}

\usepackage{microtype}

\usepackage{inconsolata}

\usepackage{graphicx}
\usepackage{xspace}
\usepackage{float}

\usepackage{enumitem}

\usepackage{fancyvrb}

\title{ATGen: A Framework for Active Text Generation}

\author{Akim Tsvigun\textsuperscript{1} \quad
Daniil Vasilev\textsuperscript{2} \quad
Ivan Tsvigun\textsuperscript{3} \quad
Ivan Lysenko\textsuperscript{2} \quad
Talgat Bektleuov\textsuperscript{1} 
\\\bf
Aleksandr Medvedev\textsuperscript{4} \quad
Uliana Vinogradova\textsuperscript{5} \quad
Nikita Severin\textsuperscript{3}\quad
Mikhail Mozikov\textsuperscript{6}\quad
\\\bf
Andrey Savchenko\textsuperscript{2,7}\quad
Rostislav Grigorev\textsuperscript{1}\quad 
Ramil Kuleev\textsuperscript{1}\quad
\\\bf
Fedor Zhdanov\textsuperscript{8}\quad
Artem Shelmanov\textsuperscript{9*} \quad
Ilya Makarov\textsuperscript{1,6*}
\\
\textsuperscript{1}Research Center of the Artificial Intelligence Institute, Innopolis University \quad
\\
\textsuperscript{2}HSE University \quad
\textsuperscript{3}Independent Researcher \quad
\textsuperscript{4}T-Technologies \quad
\textsuperscript{5}Robotics Center \quad
\\
\textsuperscript{6}AIRI \quad
\textsuperscript{7}SB-AI-Lab \quad
\textsuperscript{8}Royal Holloway University of London \quad
\textsuperscript{9}MBZUAI \\
\href{mailto:a.tsvigun@innopolis.ru}{a.tsvigun@innopolis.ru}\quad\href{mailto:artem.shelmanov@mbzuai.ac.ae}{artem.shelmanov@mbzuai.ac.ae}\quad\href{mailto:i.makarov@innopolis.ru}{i.makarov@innopolis.ru}
}

\begin{document}
\maketitle
\begin{abstract}
\input{content/abstract}

\end{abstract}

\input{content/intro}

\input{content/related_work}

\input{content/framework}

\input{content/experiments}

\input{content/conclusion}

\bibliography{custom}

\appendix

\clearpage
\onecolumn
\input{content/appendix}

\end{document}

%% file: content/abstract.tex
Active learning (AL) has demonstrated remarkable potential in reducing the annotation effort required for training machine learning models. However, despite the surging popularity of natural language generation (NLG) tasks in recent years, the application of AL to NLG has been limited. 
In this paper, we introduce Active Text Generation (ATGen) -- a comprehensive framework that bridges AL with text generation tasks, enabling the application of state-of-the-art AL strategies to NLG. Our framework simplifies AL-empowered annotation in NLG tasks using both human annotators and automatic annotation agents based on large language models (LLMs). The framework supports LLMs deployed as services, such as ChatGPT and Claude, or operated on-premises. 
Furthermore, ATGen provides a unified platform for smooth implementation and benchmarking of novel AL strategies tailored to NLG tasks. Finally, we present evaluation results for state-of-the-art AL strategies across diverse settings and multiple text generation tasks. We show that ATGen reduces both the effort of human annotators and costs associated with API calls to LLM-based annotation agents.
The code of the framework is available on GitHub\footnote{\url{https://github.com/Aktsvigun/atgen}
\\ $^*$ -- equal contribution} under the MIT license. 
The video presentation is available at \\
\url{http://atgen-video.nlpresearch.group}

%% file: content/intro.tex
\section{Introduction}

Natural language generation (NLG) has witnessed significant advancements in recent years, with the emergence of large language models (LLMs) such as o3~\cite{o3}, Claude-4-Opus~\cite{claude4}, DeepSeek-R1~\cite{deepseek-r1}, and others. These models have achieved remarkable performance across various NLG tasks, including reasoning, neural machine translation, and summarization. However, for tasks that require deep domain knowledge, such as text generation in medical or law domains, even the most powerful LLMs are not capable of generating responses of adequate quality~\cite{gpt4_medical}. Hence, for such tasks, the availability of annotated datasets still remains a critical bottleneck. Moreover, due to latency constraints and memory limitations, real-world applications often require the deployment of low-resource models. Such models often exhibit low performance without task-specific fine-tuning, further emphasizing the need for annotated data.

Recently, automatic labeling methods have been introduced to alleviate the workload of human annotators by utilizing LLMs for labeling in instruct-mode~\cite{honovich-etal-2023, wang-etal-2023-self-instruct}. Nonetheless, these techniques are not universally applicable, as current LLMs may struggle to generate high-quality annotations for domain-specific tasks and datasets. Querying the most powerful LLMs, such as o3 or Claude-4-Opus, incurs substantial costs, rendering large-scale data annotation prohibitively expensive.

Active learning (AL) is a promising approach to addressing the annotation bottleneck in machine learning. By strategically selecting for labeling the most informative instances, AL aims to maximize the model performance while minimizing the annotation effort~\cite{settles2009}. Instances are selected iteratively by batches, and after labeling each batch, the new instances are used to update an ML model, which in its turn is used to select another batch.  AL in text and token classification tasks for Transformer-based models allows reducing the number of annotations by 3-5 times compared to random selection of instances while maintaining the same level of performance~\cite{shelmanov-2021, cal-margatina}. In the era of LLM-powered annotation, AL emerges as a powerful tool -- not only streamlining human effort but also reducing the total cost of LLM API calls for automatic data labeling.

Another promising direction in this area is experimental design (ED; \citet{experimental-design-2024}). 
In this approach, instances for annotation are selected once before training the model. This helps mitigate the significant overhead in AL ushered from training a model and querying samples to label on each iteration. ED also allows for parallelizing the labeling procedure. It is especially beneficial when humans serve as annotators because it eliminates the costs associated with the latency of training a model and performing an AL query on each iteration. However, to select new instances, ED utilizes neither labels obtained during the annotation process nor, consequently, the model knowledge after fine-tuning on the already annotated instances. This can potentially degrade its benefits compared to those of AL. For the sake of simplicity, for the remainder of the paper, we subsume ED approaches under the umbrella term AL since ED can be considered a particular case of AL.

Although there exist many NLP-oriented AL frameworks, they primarily focus on classification and sequence labeling tasks~\cite{alpacatag, altoolbox, schroder-etal-2023-small}. Furthermore, launching an AL cycle with modern LLMs requires parameter-efficient tools for fine-tuning (PEFT) and support for efficient LLM inference. Despite the recent progress in AL strategies for NLG tasks~\cite{idds, hadas-2024, huds-2024}, there is currently no unified framework to evaluate these strategies in unified settings. Finally, given the remarkable performance of modern LLMs, powerful models can often effectively replace human annotators for labeling data on many relatively simple tasks~\cite{fabricator, peng2023instructiontuninggpt4}.

To fill the aforementioned gaps, we introduce Active Text Generation (ATGen) -- a comprehensive framework that enables AL annotation in NLG tasks. ATGen aims to democratize active learning for text generation by making it accessible to users regardless of their expertise level in these topics. With just a few lines of code, users can initiate AL-empowered data annotation with human or LLM-based annotators. For researchers, the framework offers a unified platform for developing and benchmarking novel active learning strategies, thereby fostering further innovation in the field.

The main contributions of our framework can be summarized as follows:
\begin{itemize}
    \item A collection of state-of-the-art AL and ED strategies for text generation implemented with unified program interfaces.
    \item A demo web application that allows performing AL annotation for NLG tasks, supporting both manual labeling and automatic labeling via proprietary or open-source LLM-based annotation agents.
    \item A benchmarking platform for rigorous evaluation of AL strategies in NLG tasks.
    \item Experimental evaluation demonstrating that AL substantially reduces annotation time for manual labeling and total costs for LLM API calls in automatic annotation scenarios.
\end{itemize}

%% file: content/related_work.tex
\section{Related Work}

\subsection{AL Selection Strategies in NLP}

\paragraph{Non-generative NLP tasks.}
AL has been widely explored for non-generative NLP tasks~\cite{yuan-etal-2020-cold, cal-margatina, shelmanov-2021, liu-2021, tsvigun-etal-2022-towards}. Most prominent solutions substantially outperform random sampling, emphasizing the benefits of AL, as shown by~\citet{schroder-etal-2022-revisiting}. Particularly, for text  classification and token classification tasks, enabling AL allows reaching the same model quality with a 3-6 times reduced budget for annotation.

Technically, some of these strategies can be reused for generative tasks; however, their performance in these tasks is questionable. For example, least confidence~\cite{lewis_lc}, prediction entropy~\cite{roy_entropy}, and Coreset strategies~\cite{coreset-2018} were shown to substantially outperform random sampling in text classification~\cite{schroder-etal-2022-revisiting, altoolbox}. However, their extension in generative tasks does not improve the quality obtained through random sampling and can even lead to lower results~\cite{idds, al-nlg-2023}. Therefore, application of such strategies to NLG tasks requires careful evaluation in various settings before usage.

\paragraph{Text generation tasks.}
~\label{al_nlg_strategies}
Recently, several AL strategies tailored to NLG tasks have emerged. \textbf{TE-delfy}~\cite{te-delfy-2020}, introduced for the NMT task, combines uncertainty-based token entropy (TE) and model-agnostic decay logarithm frequency (delfy). \textbf{BLEUVar}, proposed by \citet{bleuvar}, considers documents as vectors and employs the BLEU score~\cite{bleu} to compute the dissimilarity between them. It selects for annotation texts that exhibit the highest variability in BLEU scores across multiple stochastic generations. \textbf{NGF-SMP}~\cite{hu-neubig-2021-phrase} selects frequent phrases that are underrepresented in the labeled data. \textbf{HUDS}~\cite{huds-2024}
combines uncertainty-based and metric learning approaches by using normalized negative log-likelihood to estimate uncertainty for unlabeled sentences and stratifying the data based on these scores.
The final score is a weighted sum of the uncertainty score and the cosine distance between the sentence's BERT embedding and its corresponding stratum centroid embedding.
\textbf{HADAS}~\cite{hadas-2024}, introduced for Abstractive Text Summarization (ATS), assesses the susceptibility of a generative model to hallucinations across semantic frame, discourse, and content verifiability errors. It combines entailment-based semantic frame scoring, QA-based discourse evaluation, and BERTScore-based content verifiability assessment to produce a hallucination-aware score for each text.
\textbf{LDCAL}~\cite{ldcal}, also designed for the ATS task, fuses curriculum learning and active learning by leveraging an LLM-determined difficulty score to partition documents into four levels and then selecting representative instances that maximize the model’s certainty gain, thus covering both high- and low-density regions.

Some works favor ED approaches since the query phase of AL in NLG can incur significant overhead, especially when the model is required to generate some text to assess the informativeness of the instance. \textbf{IDDS}~\cite{idds} selects instances with low semantic similarity with the labeled pool and high similarity with the whole unannotated pool. \citet{facility-location-2024} suggest using traditional submodular functions for subset selection. They demonstrate the effectiveness of the facility location function~\cite{francis1991discrete} in some settings. We will refer to this strategy as ``Facility Location'' throughout the paper.

\subsection{Existing AL Frameworks for NLP}

There exist many libraries that allow running AL for various NLP tasks~\cite{klie-etal-2018-inception, alpacatag, nguyen-etal-2022-famie, altoolbox,  schroder-etal-2023-small}. However, these systems miss many practical features and tools, crucial for seamless integration with data analysis pipelines and annotation. 
Features essential for seamless data annotation in text generation tasks with active learning integration are predominantly scattered across various frameworks. 
\citet{Huang2021deepal, distil2021} implement many state-of-the-art strategies for classification tasks. \citet{schroder-etal-2023-small} provides unified interfaces for benchmarking AL on text classification datasets.
ALToolbox \cite{altoolbox} provides a pre-implemented set of AL strategies and a GUI for text annotation tasks such as text classification and information extraction. It also allows benchmarking AL strategies for encoder-based and sequence-to-sequence models. The tool proposed by \citet{fabricator, adala} allows annotating data using LLMs, but has no integration with AL. Finally, Argilla~\cite{argilla} offers a comprehensive tool for data annotation and quality improvement in AI projects, but also lacks AL workflow support.

Additionally, running an AL loop with modern LLMs is both time- and memory-consuming and requires enabling approaches for efficient fine-tuning and inference. To our knowledge, ATGen is the first framework to synergize PEFT approaches like LoRA and inference-efficient frameworks like vLLM~\cite{paged-attn} for efficient AL.

%% file: content/framework.tex
\section{ATGen Description}

In AL, one starts with a small labeled dataset and a large pool of unlabeled data. An acquisition model is trained on the labeled set, then used to evaluate the unlabeled data. A selection AL strategy is used to identify the most informative instances, which are then labeled by an oracle (a human or a model). This process repeats iteratively, gradually improving the model's performance until some stopping criteria are met.

ATGen supports all stages of AL in NLG. It includes: (1) a web application for manual annotation with integrated AL support; (2) automatic AL-guided data annotation using LLMs, optimized for cost-efficient API usage; (3) a wide range of implemented AL query strategies, evaluation metrics, and configurable stopping criteria; (4) tools for efficient model fine-tuning and inference; (5) a user-friendly web interface for configuration and monitoring; and (6) benchmarking scripts for evaluating and comparing AL strategies across tasks and domains.

\subsection{Framework Features}

\input{figures/web}

\subsubsection{AL Strategies for NLG Tasks}

ATGen implements all the state-of-the-art AL selection strategies in NLG (see  Section~\ref{al_nlg_strategies}).
We also incorporate various uncertainty-based strategies, such as normalized sequence probability~\cite{nsp-ueffing}, mean token entropy~\cite{te-delfy-2020}, and others. 

\subsubsection{Web GUI for Manual Labeling}

ATGen can be used for manual text annotation via a web GUI application.
For manual annotation, we recommend using ED strategies, as they require only a single round of annotation before training the target model. This approach significantly reduces annotation time and delays, making it especially suitable for scenarios involving human annotators. The GUI for annotation is displayed in Figure~\ref{fig:annotation}.

\input{figures/annotation}

\subsubsection{Automatic Labeling using LLMs}

Users can select any chat-based model from a range of providers to serve as an annotator in place of a human. ATGen integrates seamlessly with leading API-based LLM providers, including \textbf{OpenAI}, \textbf{Anthropic}, and other OpenAI-compatible platforms such as Nebius AI Studio.  For optimal annotation quality, we suggest using Claude-3.5-sonnet or GPT-4o models. For OpenAI-based models, we implement the batched API, which is 50\% cheaper and several times faster compared to its synchronous analogue.
Users can also choose a model from HuggingFace or supply a custom model for data annotation. The selected model runs locally on the user’s hardware and processes the input data in batch mode.

\subsubsection{Supported Acquisition Models and Datasets}
The framework is tightly integrated with HuggingFace. It supports any acquisition model available through the HuggingFace Transformers library~\cite{Wolf2019HuggingFacesTS} and
allows pulling data from the HuggingFace hub via the datasets library~\cite{lhoest-etal-2021-datasets}. Users can also upload their own datasets in either CSV or JSON format.

\subsubsection{Efficient Fine-Tuning and Inference}

Most AL strategies require fine-tuning and/or inference with the target model. Since modern LLMs have billions of parameters, the implementation of computationally efficient methods for training and inference becomes crucial for real-world applications of the framework. ATGen, therefore, supports several parameter-efficient fine-tuning methods \cite{houlsby2019parameter}: LoRA~\cite{lora-2022}, which approximates the update matrix as the product of two low-rank matrices; QLoRA~\cite{qlora-2023}, which further reduces memory usage using the 4-bit NormalFloat data type, double quantization, and paged optimizer; and DoRA~\cite{dora-2024}, which allows for more expressive fine-tuning by introducing an additional low-rank matrix to model both additive and multiplicative updates. The user can omit the usage of PEFT; yet, this will require a large amount of GPU memory.

For efficient inference, ATGen leverages three modern inference-accelerating frameworks: \textbf{vLLM}, which optimizes the inference through PagedAttention~\cite{paged-attn}, \textbf{SGLang}~\cite{sglang}, which leverages RadixAttention for prefix caching along with other techniques, and \textbf{Unsloth}~\cite{unsloth}, which accelerates the inference through various optimizations like memory-efficient kernels.

\subsubsection{Supported Evaluation Metrics}

Performance evaluation in NLG tasks is a crucial bottleneck, since there are many perspectives from which the quality of the generation can be gauged~\cite{NEURIPS2021_e4d2b6e6}. We split the implemented metrics into three groups:
\begin{enumerate}[itemsep=0pt, parsep=0pt]
    \item Automated metrics. These are traditional metrics such as BLEU~\cite{bleu}, ROUGE~\cite{rouge}, and others, aimed at automatic evaluation of results.
    \item Open-source LLM-based metrics. This group incorporates metrics that require the usage of an open-source model, such as BERTScore~\cite{bert-score}, BARTScore~\cite{NEURIPS2021_e4d2b6e6}, AlignScore~\cite{alignscore}, and others.
    \item Proprietary LLM-based metrics. Recently, quality estimation with LLMs gained much attention and has been adopted in many works~\cite{liu-etal-2023-g}. We therefore use the DeepEval~\cite{deepeval} framework for evaluation via LLMs. We note that this type of evaluation incurs additional computational cost compared to other methods. Therefore, we recommend using it only at the final stage of active learning for the ultimate assessment of model performance.
\end{enumerate}

\subsection{Demo Web Application}

ATGen allows a user to deploy a web application for AL annotation on the user's dataset with either human or LLM serving as an annotator in just one line of code. To launch AL annotation from the GUI, a user can configure a labeler, the data for annotation, and a stopping criterion from a web form (Figure~\ref{fig:web}). 
The application supports several stopping criteria: annotating a fixed number of texts, reaching a certain level of the model's performance on a test set, or running out of budget when using a human or an API-based LLM agent for annotation.

There are many other parameters when running AL: training hyperparameters, PEFT-related parameters, and others. To customize additional parameters, a user can create a configuration file and apply it using the submission form in the top left corner.

After each AL iteration, the model performance is evaluated either on the test data, if available, or on the test split of the labeled data.

\subsection{Benchmark for AL Selection Strategies}

ATGen provides benchmarking scripts for a seamless evaluation of the performance of AL strategies in NLG tasks. Running an experiment requires implementing the custom strategy according to the guidelines. The benchmarking tool can also be leveraged to test the existing approaches in various AL settings (e.g. in new domains, with LLM annotators, etc.). 
The example code to benchmark a strategy is provided in Figure~\ref{fig:benchmarking_code_example} in Appendix~\ref{app:example_code}.

%% file: figures/web.tex
\begin{figure}[t]
    \vspace{-0.1cm}
	\center{\includegraphics[width=1.\linewidth]{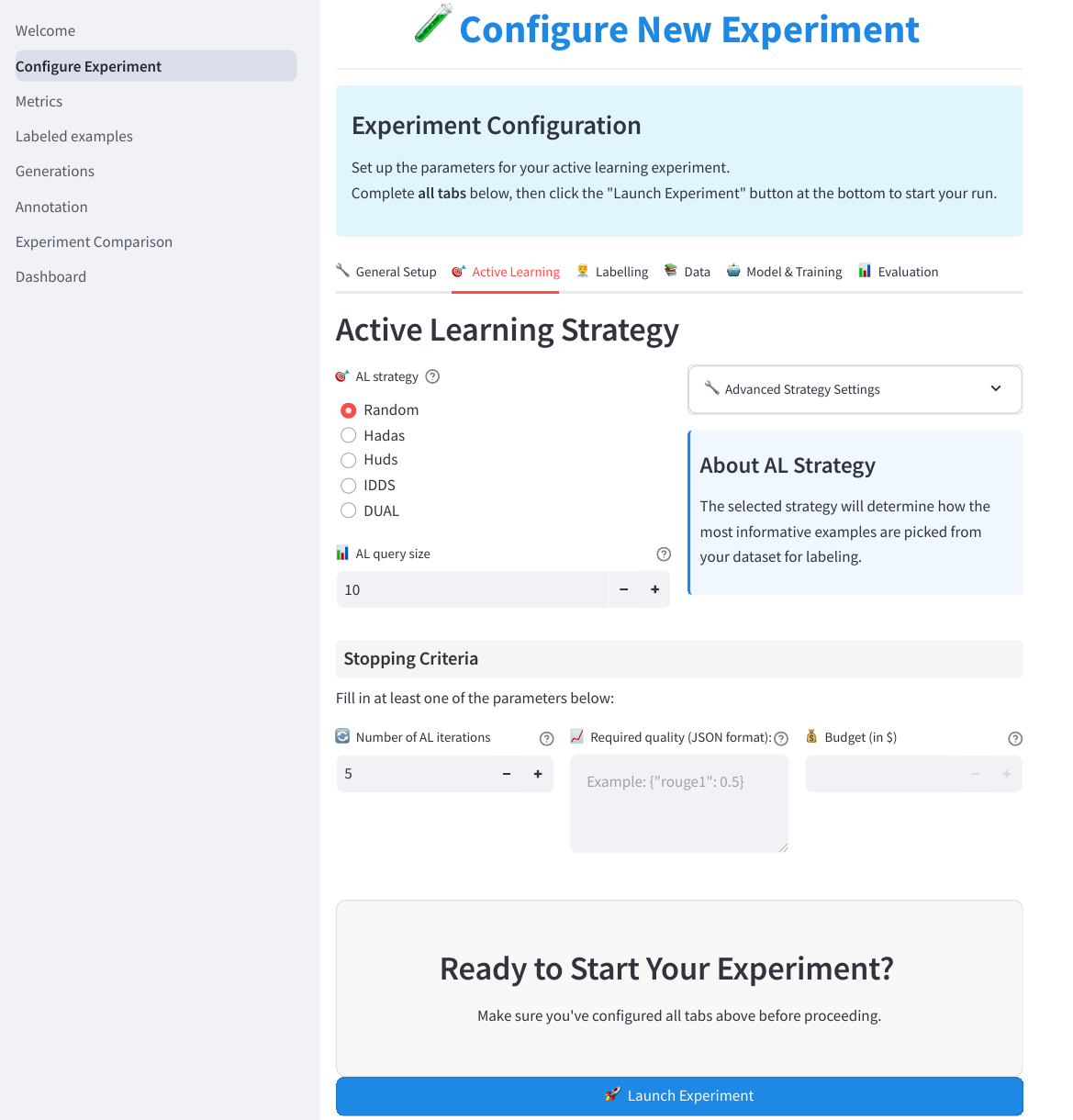}}
	\vspace{-0.7cm}
	\caption{The ATGen configuration form to launch active learning in a GUI.}
	\label{fig:web}
	\vspace{-0.3cm}
	
\end{figure}

%% file: figures/annotation.tex
\begin{figure}[h]
    \vspace{-0.1cm}
	\center{\includegraphics[width=1.\linewidth]{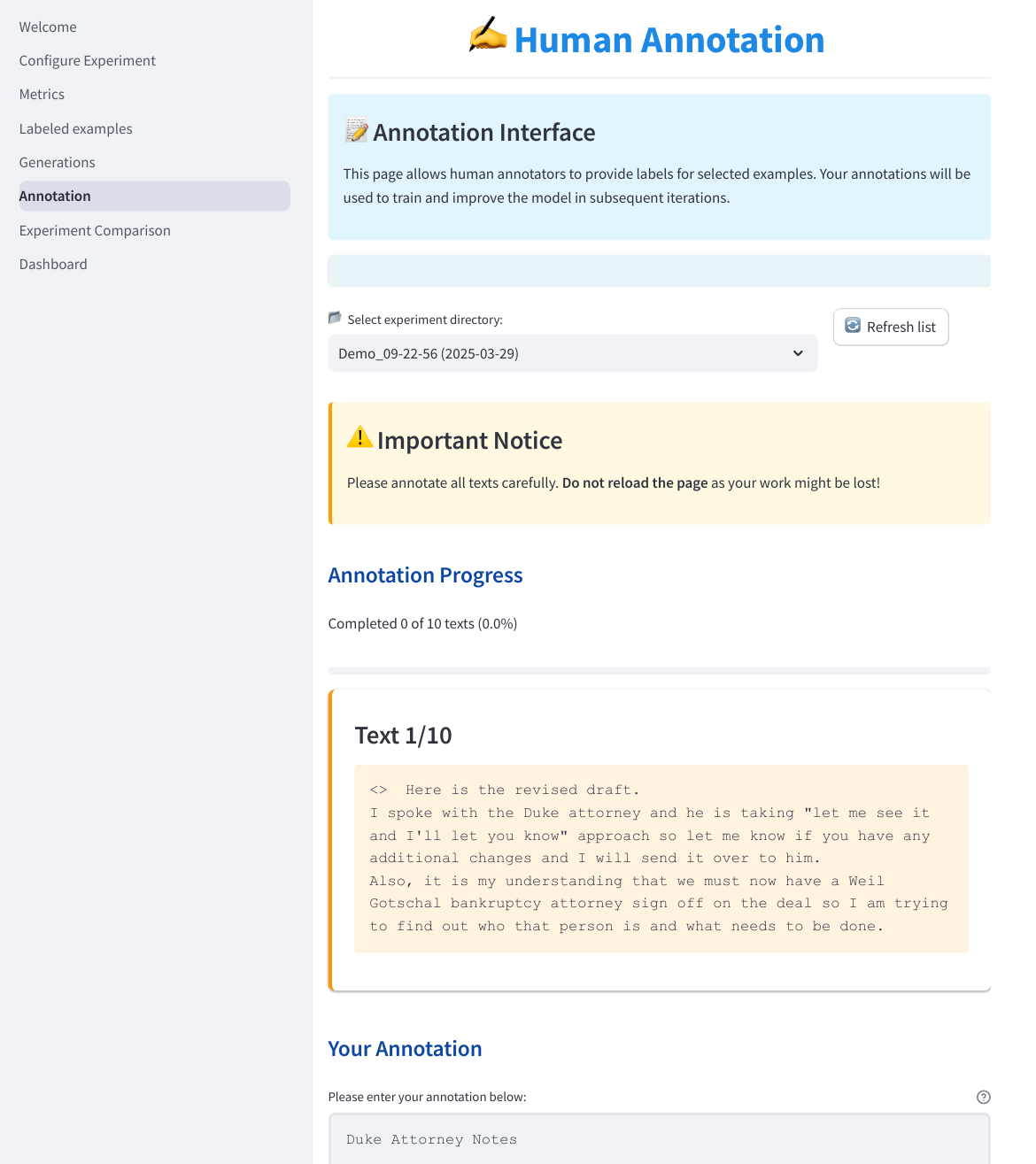}}
	\vspace{-0.7cm}
	\caption{Human annotation page interface.}
	\label{fig:annotation}
	\vspace{-0.3cm}
	
\end{figure}

%% file: content/experiments.tex
\input{figures/triviaqa}

\section{Experiments}

Using the ATGen benchmark, we evaluate the performance of AL and ED methods in various setups.

\subsection{Experimental Setup}

\subsubsection{AL Settings} 

We adopt the widely used simulated AL experimental setup~\cite{settles-craven-2008-analysis,shen2018,idds}, which emulates the AL annotation cycle. 
At each iteration, we select a fixed number of top-ranked instances from the unlabeled pool according to the AL query strategy and assign them their ground-truth outputs, simulating an annotation by an oracle.

The selected instances are removed from the unlabeled pool and added to the training dataset for subsequent iterations. We then train a new acquisition model from the previous checkpoint using the accumulated training data and evaluate its performance on the test set. This process produces a curve that illustrates how model performance depends on the amount of annotated training data. A higher curve indicates better performance of the AL query strategy, as it reflects the model's ability to achieve better results with less training data.
For robustness, we run each experiment several times with different random seeds and average the obtained curves.

Given the growing interest in LLMs application for data labeling~\cite{honovich-etal-2023}, we also conduct experiments where an LLM annotates the data instead of using ground-truth annotations. In this scenario, we use DeepSeek-R1 as the annotation model.

\subsubsection{Datasets, Metrics, and Acquisition Model}

Following the recent works in this area~\cite{idds, hadas-2024}, we evaluate AL and ED strategies on 4 diverse NLG tasks: open-domain question answering: TriviaQA~\cite{triviaqa}, Wiki subset; mathematical reasoning: GSM8K~\cite{gsm8k}; reading comprehension: RACE~\cite{race}; and text summarization: AESLC~\cite{aeslc}. For TriviaQA and GSM8K, we select 1\% of texts from the train set for labeling on each AL iteration. For RACE and AESLC, we select 10 texts to label on each AL iteration to align with the previous works.

Due to space limitations, we present results for the TriviaQA and GSM8K datasets in the main part of the paper, while results for RACE and AESLC are provided in Appendix~\ref{app:exps}.

We perform experiments with an emulation of manual labeling on all datasets.
On TriviaQA and GSM8K, we also conduct experiments with LLM-based labeling by DeepSeek-R1.

We run the experiments with the Qwen/Qwen3-1.7B acquisition model\footnote{\url{https://huggingface.co/Qwen/Qwen3-1.7B}} -- one of the state-of-the-art models to date in its size. The hyperparameters are presented in Appendix~\ref{app:hyperparams}.

To assess the performance of the acquisition model, we use the exact match (EM) metric for GSM8K and RACE, the relaxed version of EM that accepts any of the valid answers for TriviaQA, and ROUGE-2~\cite{rouge} along with AlignScore~\cite{alignscore} for AESLC to ensure that the increased ROUGE score is not caused by an increase in hallucinations. 

\subsection{Results}

\input{figures/gsm8k}

Figure~\ref{fig:triviaqa} presents the performance of AL query strategies on the TriviaQA dataset under both manual annotation emulation and LLM-based annotation scenarios. The results reveal consistent patterns across both settings, with HUDS, HADAS, and Facility Location strategies substantially outperforming random sampling across all iterations. 

For example, random sampling requires over 12\% of the dataset to be labeled to match the performance level that AL achieves with three times less data (just 4\%) -- in both the ``manual'' and LLM-based annotation scenarios.

Figure~\ref{fig:gsm8k} shows analogous experiments on the GSM8K dataset. Under manual annotation emulation (Figure~\ref{fig:gsm8k}a), the same three strategies plus IDDS demonstrate superior performance compared to random sampling throughout all iterations. However, when using LLM-based annotation (Figure~\ref{fig:gsm8k}b), we observe a significant degradation in overall quality across all strategies. While HUDS, HADAS, and Facility Location maintain the advantage over random sampling, absolute quality scores decrease by several percentage points. The performance gap likely stems from the inherent limitations of the oracle, DeepSeek-R1. Despite being a state-of-the-art LLM on mathematical reasoning tasks, it is still prone to making annotation errors that accumulate through the AL process. This finding underscores that for specialized domains, human annotation remains crucial for obtaining high-quality models.

Additional experiments on RACE and AESLC datasets (Figures~\ref{fig:race} and \ref{fig:aeslc} in Appendix~\ref{app:exps}) corroborate these findings. HUDS, HADAS, and Facility Location consistently outperform random sampling across diverse NLG tasks. These strategies substantially reduce annotation costs by achieving target quality levels with far fewer labeled examples.

Overall, the results demonstrate that AL works effectively in both manual annotation and LLM-based annotation scenarios. This means that AL can reduce costs for LLM API calls by 2-4x when AL annotation is executed in a fully automatic regime, while achieving the same level of performance. Although AL requires retraining a small LLM on each iteration, which consumes computational resources, this process can be executed on a user's hardware, making it effectively ``free'' for the user. Therefore, despite the additional computational expenses, the savings on LLM API calls are significantly more substantial.

%% file: figures/triviaqa.tex
\begin{figure*}[h!]
    \footnotesize
    
    \centering
    \begin{minipage}[h]{0.49\linewidth}
    \center{\includegraphics[width=1\linewidth]{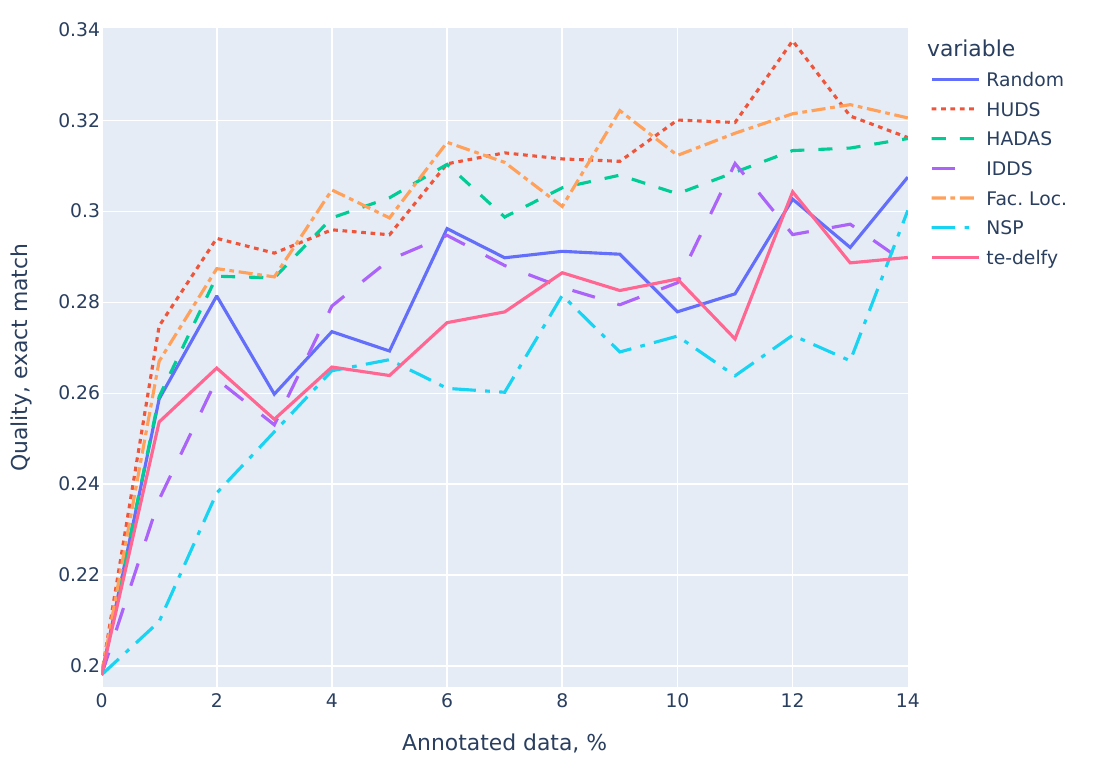} a) TriviaQA dataset with emulation of ``manual'' labeling.}
    \end{minipage}
    \hspace{0.1cm}
    \begin{minipage}[h]{0.49\linewidth}
    \center{\includegraphics[width=1\linewidth]{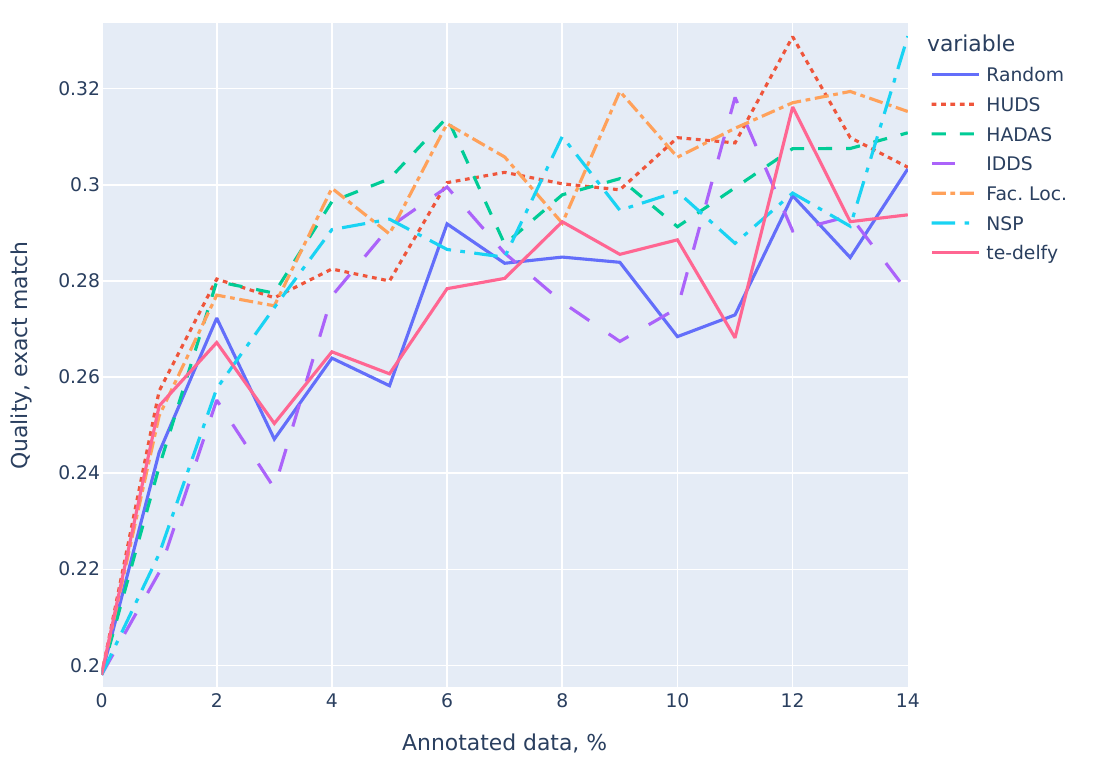} b) TriviaQA dataset with labeling using an LLM.}
    \end{minipage}
    
    \caption{Performance of AL selection strategies on the TriviaQA dataset with different annotation sources.}
    \label{fig:triviaqa}
\end{figure*}

%% file: figures/gsm8k.tex
\begin{figure*}[h!]
    \footnotesize
    
    \centering
    \begin{minipage}[h]{0.49\linewidth}
    \center{\includegraphics[width=1\linewidth]{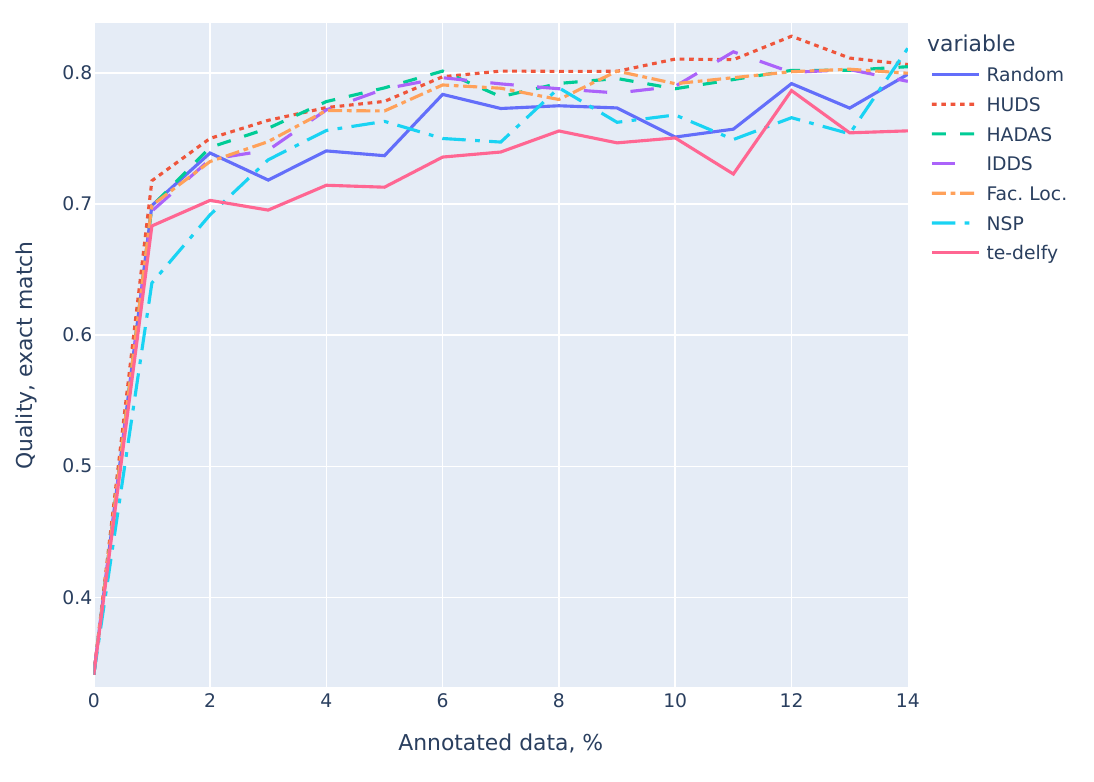} a) GSM8K dataset with emulation of ``manual'' labeling.}
    \end{minipage}
    \hspace{0.1cm}
    \begin{minipage}[h]{0.49\linewidth}
    \center{\includegraphics[width=1\linewidth]{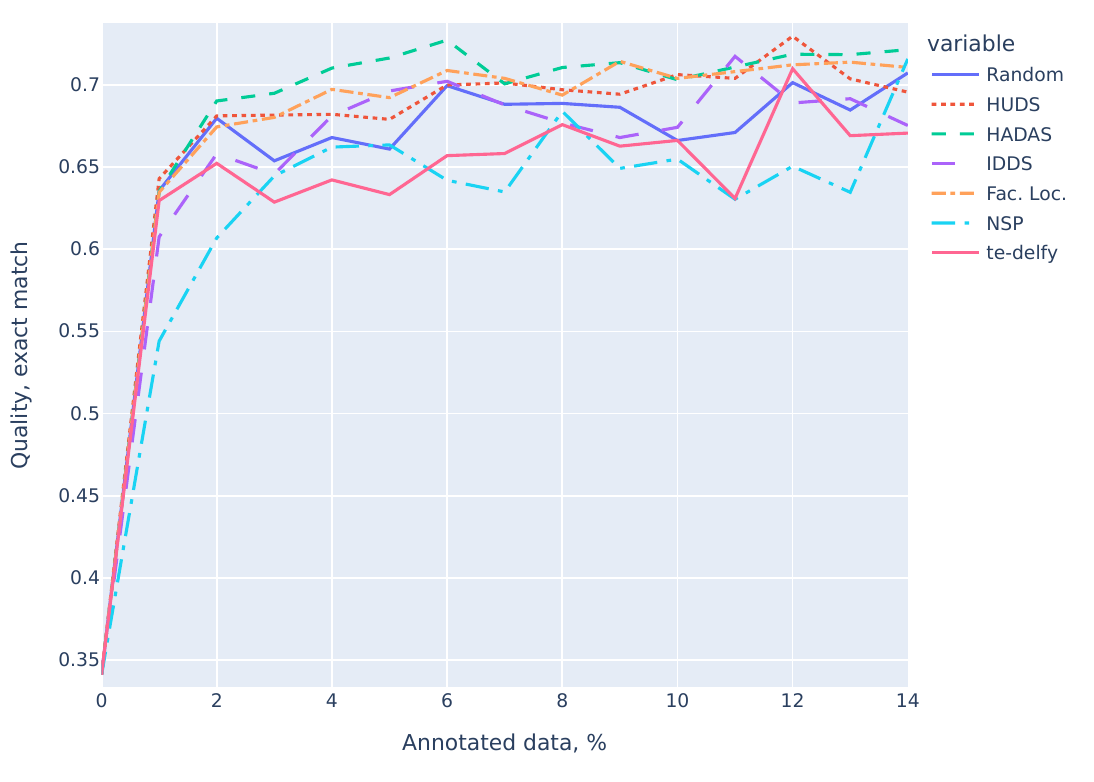} b) GSM8K dataset with labeling using an LLM.}
    \end{minipage}
    
    \caption{Performance of AL selection strategies on the GSM8K dataset with different annotation oracles.}
    \label{fig:gsm8k}
\end{figure*}

%% file: content/conclusion.tex
\section{Conclusion}

We have presented ATGen -- a comprehensive framework for AL in text generation tasks. The framework implements all state-of-the-art active learning techniques for NLG, offers a web-based annotation tool that supports both human and LLM-assisted labeling, and includes scripts for consistent experimental evaluation of AL query strategies. We have also conducted experiments, demonstrating that state-of-the-art strategies like HUDS can save annotators' time and budget for LLM API calls.

We believe that AL is a promising approach even in the era of powerful LLMs, as it can help to reduce costs for building smaller models that could be deployed in production. We hope that our framework will foster the development of better AL strategies in the future.

\section*{Limitations}

We have not investigated possible bias introduced by active learning during annotation. This is an important future work as AL strategies might alter the data distribution significantly.

We note that AL requires some additional computational expenses for re-training the target LLM. If the target LLM is not big, these expenses might be negligible. However, for bigger LLMs, that might be an additional concern.

\section*{Ethical Considerations}

For experiments and demo implementation, we reused pre-existing corpora and LLMs, which have been publicly released and approved for research purposes. The code of the demo has been released under the MIT license on \href{https://github.com/Aktsvigun/atgen}{GitHub}.

Using LLMs for automatic annotation should be approached with caution, as these models inherit social biases and often produce hallucinations. Hence, additional verification of annotation quality is required.

\section*{Acknowledgements}

The work of A.~Tsvigun, T.~Bekleutov, R.~Grigorev, R.~Kuleev, and I.~Makarov on Sections 2--4 was supported by the Research Center of the Artificial Intelligence Institute at Innopolis University and financially by the Ministry of Economic Development of the RF (code 25-139-66879-1-0003).

%% file: content/appendix.tex
\section{Additional Experiments}~\label{app:exps}

\input{figures/aeslc}
\input{figures/race}

\newpage
\section{Model Hyperparameters for Benchmarks}~\label{app:hyperparams}
\input{figures/hyperparameters}

\section{Code Examples}~\label{app:example_code}
\input{figures/benchmarking_code}

%% file: figures/aeslc.tex
\begin{figure*}[!h]
    \footnotesize
    
    \centering
    \begin{minipage}[h]{0.49\linewidth}
    \center{\includegraphics[width=1\linewidth]{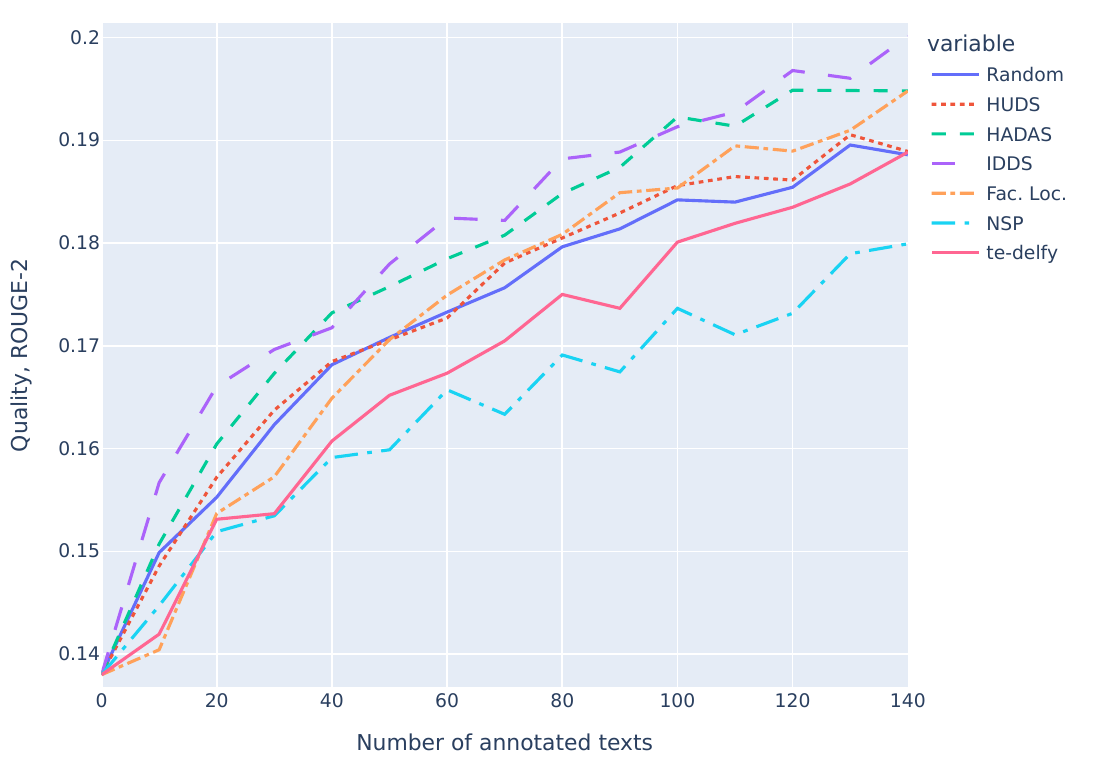} a) ROUGE-2 scores.}
    \end{minipage}
    \hspace{0.1cm}
    \begin{minipage}[h]{0.49\linewidth}
    \center{\includegraphics[width=1\linewidth]{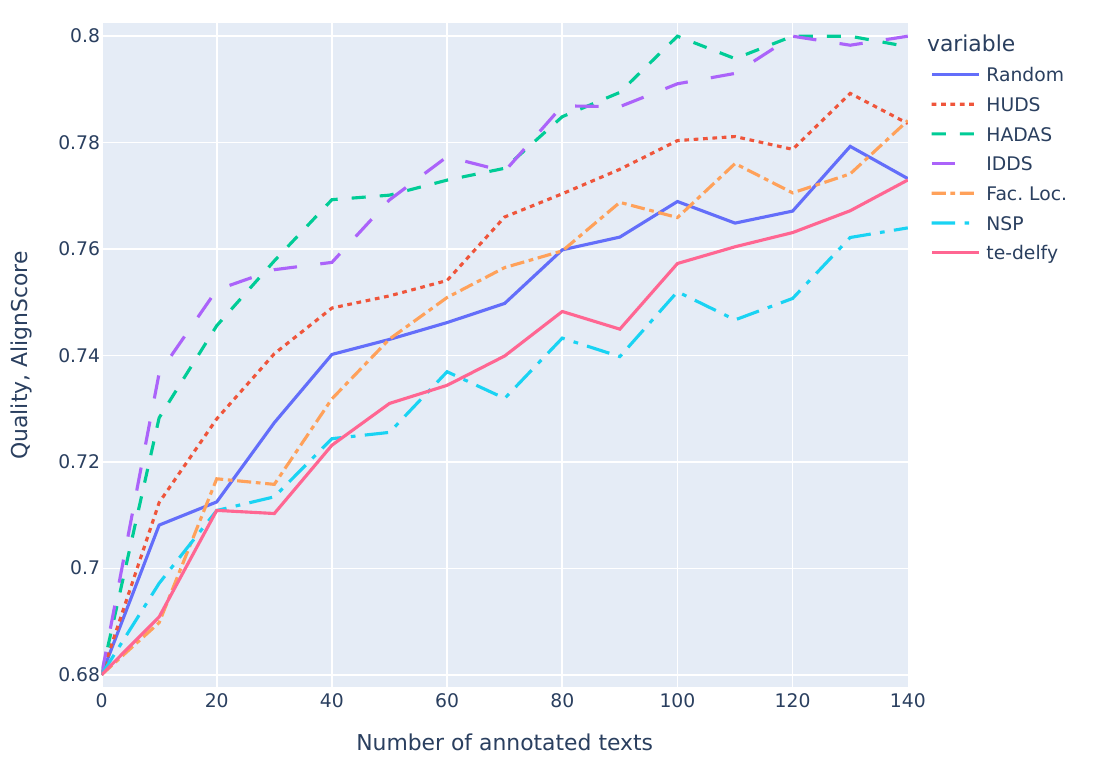} b) AlignScore scores.}
    \end{minipage}
    
    \caption{Performance of AL and ED strategies with emulation of ``manual'' labeling on AESLC in terms of the main metric (ROUGE-2) and a hallucination-sensitive metric (AlignScore).}
    \label{fig:aeslc}
\end{figure*}

%% file: figures/race.tex
\begin{figure*}[!h]
    \footnotesize
    
    \centering
    \includegraphics[width=0.49\linewidth]{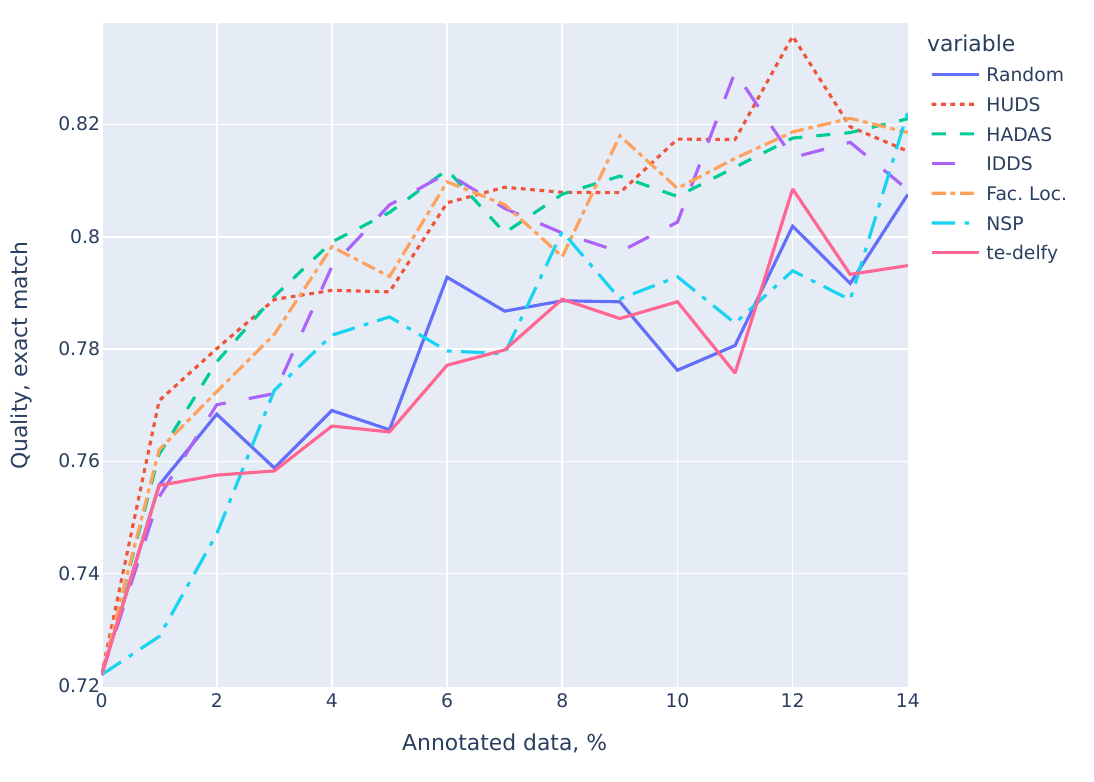}
    
    \caption{Performance of AL and ED strategies with emulation of ``manual'' labeling on the RACE dataset.}
    \label{fig:race}
\end{figure*}

%% file: figures/hyperparameters.tex
\begin{table}[ht!]
\centering
\footnotesize
\label{tab:hyperparams}

{\begin{tabular}{lll}
\textbf{Hparam} & \textbf{Value} \\ \hline
Checkpoint & Qwen/Qwen3-1.7B \\
\# Param. & 1.7B \\
Quantization & None \\
\hline
Number of epochs & 5 \\
Train Batch size & 16 \\
Evaluation Batch size & 16 \\
Evaluation Split Size & 20\% \\
Gradient Accumulation Steps & 1 \\
Learning Rate & 3e-5 \\
Warmup Ratio & 0.03 \\
Weight Decay & 0.01 \\
Max. Gradient Norm & 1 \\
Early Stopping Patience & 5 \\
Optimizer & adamw\_hf \\
\hline
Inference Framework & vLLM \\
Batch Size & 16 \\
GPU Memory Utilization & 0.5 \\
Temperature & 0 \\
Generation \verb|top_p| & 0.5 \\
\hline
PEFT & Enabled \\
\verb|r| & 16 \\
\verb|lora_alpha| & 16 \\
\verb|lora_dropout| & 0. \\
LoRA \verb|bias| & \verb|'none'| \\
\hline
\end{tabular}}
\caption{Hyperparameter values and checkpoints from the Hugging Face repository~\cite{Wolf2019HuggingFacesTS} of the models.}
\end{table}

%% file: figures/benchmarking_code.tex
\begin{figure*}[!h]
    \footnotesize
    
    \centering
    \begin{BVerbatim}
HYDRA_CONFIG_NAME=base python scripts/run_active_learning.py al=STRATEGY_NAME
    \end{BVerbatim}
    
    \caption{A Bash command example to benchmark a AL strategy with the name ``STRATEGY\_NAME''.}
    \label{fig:benchmarking_code_example}
\end{figure*}

\begin{figure*}[!h]
    \footnotesize
    
    \centering
    \begin{BVerbatim}
HYDRA_CONFIG_NAME=base run-al \
    data=gsm8k \
    al.init_query_size=0.01 \
    al.query_size=0.01 \
    al.num_iterations=20 \
    al=huds \
    evaluation.additional_metrics=[] \
    labeller=api_llm \
    labeller.parameters.model=gpt-4.1 \
    labeller.parameters.max_tokens=4096 \
    al.budget=100 \
    labeller.price.input_per_1m=2 \
    labeller.price.output_per_1m=8 \
    labeller.api_key=<your_openai_api_key>
    \end{BVerbatim}
    
    \caption{Advanced Bash code example to benchmark the strategy ``huds'' on the dataset ``TriviaQA'', annotating 1\% of texts on each iteration, with GPT-4.1 LLM serving as labeller, calculating only the 'relaxed' exact match metric.}
    \label{fig:benchmarking_code_example_advanced}
\end{figure*}